\documentclass[submission]{IEEEtran}

\usepackage{cite}      

\usepackage{graphicx}  

\usepackage{subfigure} 

\usepackage{url}       

\usepackage{amsmath}   

\usepackage{widetext}
\usepackage[section]{placeins}
\usepackage{fancyhdr}
\usepackage{listings}
\usepackage{color}
\usepackage{epsfig}
\usepackage{algorithmic,algorithm,cite,endnotes,balance,amssymb}
\floatstyle{ruled}
\newfloat{algo}{thp}{lop}
\floatname{algo}{Algorithm}

\begin{document}

\title{An IMU-Aided Carrier-Phase Differential GPS Positioning System}
\author{Shuqing~Zeng,~\IEEEmembership{Member, IEEE}
\thanks{Manuscript was drafted on May 16, 2012
}
}
%
%
%
\markboth{An IMU-Aided Carrier-Phase Differential GPS Positioning System}{S. Zeng}
%



\maketitle

\begin{abstract}
We consider the problem of carrier-phase differential GPS positioning for an land vehicle navigation system (LVNS), tightly coupled with an inertial measurement unit (IMU) and a speedometer. The primary focus is to apply Bayesian network to an IMU-aided GPS positioning system based on carrier-phase differential GPS. We describe the implementation details of the positioning system that integrates GPS measurements (i.e., pseudo-range, carrier-phase and doppler), IMU measurements, and speedometer measurements. We derive the linearized state process equation and the measurement equation for GPS and speedometer. To account for constraints of land vehicle, we add two more pseudo measurements to ensure the perpendicular velocities close to zero.
\end{abstract}

\begin{IEEEkeywords}
Differential Carrier-phase GPS, Land Vehicle Navigation System, IMU aided GNSS, Bayesian Network
\end{IEEEkeywords}

\section{INTRODUCTION}
Global navigation satellite system such as GPS based positioning systems are in widespread use world-wide. It is possible to determine the position as accurate as a few centimeters if a differential configuration using a fixed known base station is applied. However this GPS system configure requires line-of-sight to the satellites. In urban areas with high buildings or in forests, the quality of the position estimate degrades due to multi-path effects or even leads to a signal outage (e.g., in tunnels or under the bridges). Another drawback of GPS based system is the slow update rate of GPS measurements. For applications such as autonomous driving, a more frequent estimation of vehicle position, velocity, and attitude is required.

Inertial measurement unit (IMU) can provide such desired information for autonomous driving. Using accelerometers and gyroscopes, and Newton's law of motion, IMU can determine the position, velocity, and attitude of the vehicle. IMU is a self-contained sensor and provides inertial measurement at a higher rate (e.g., 100 Hz for consumer grade devices). Since IMU measures the relative increment from the previous known state, a integration process (call dead-reckoning) is needed. Because of this integration, errors caused by sensor bias, sensor scale factor, and sensor nonlinearity are accumulated, and may yield unbounded drifts of the position and attitude estimation of the vehicle.

Fusion systems integrating GPS with global accuracy and an IMU with local accuracy becomes the mainstream technology for land vehicle navigation system (LVNS) \cite{Wagner2005}. The GPS measurement aids the integration such that the drifting errors are bounded and, on the other hand, the IMU measurement can be used to estimate GPS carrier phase cycle, and identify and correct cycle estimation error when cycle slip occurs.

In this paper the primary focus is to apply the Bayesian network (BN) proposed in \cite{Zeng_TIM2009} to an IMU-aided GPS positioning system based on carrier-phase differential GPS. We describe the implementation details of the positioning system that integrates GPS measurements (i.e., pseudo-range, carrier-phase and doppler), IMU measurements, and speedometer measurements. We derive the linearized state process equation to express the evolution of the augmented vehicle state consisting of vehicular position, velocity, attitude, and error parameters of IMU measurement (e.g., bias and scale factor). Also the measurement equation for GPS measurement is derived in term of the augmented state vector. To account for land vehicle that does not slip and travels along the bore-sight, we add two more pseudo measurements to ensure the perpendicular velocities are close to zero.

Integration of GPS and IMU is a well-studied area \cite{Farrell2000,Godha2007,MacGougan2010,Soon2008,Georgy2011,Li2011-GPSSolut} and successfully used in practice~\cite{Waegli2009,Kennedy2008}. Due to the fact that LVNS typically has to operate in areas where GPS signals are either blocked or severely degraded, ambiguity resolution (AR) of double-difference carrier phase data as integers is still a challenge problem. A few tens of seconds of data is required for AR to converge to a correct solution. However, the time between two consecutive dropouts for satellite may be much shorter than this requirement duration. Therefore, the AR process may be prematurely terminated due to outages, and new AR ones need to be started on-the-fly when satellites arise in the GPS-adverse environment.

The rest of this paper is organized as follows. Section \ref{SC:IMUprocessing} is devoted to the details of IMU data processing. Section \ref{SC:GPSprocessing} is focused on GPS data processing. Section \ref{SC:SensorErrorModel} outlines the stochastic model of the sensor errors. In Section \ref{SC:IMU_GPS_Integration} we discuss the algorithm to integrate data from IMU, GPS, and vehicle speedometer for positioning and attitude estimation of land vehicle. Finally we give concluding remarks in Section \ref{SC:Conclusion}.

\section{IMU Data Processing} \label{SC:IMUprocessing}
\subsection{Coordinate Frames}
\begin{figure}[h]
    \centering
    \includegraphics[width=6cm]{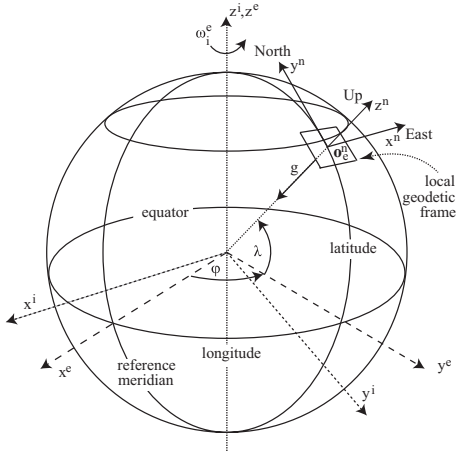}
    \caption{\protect\small  Earth-centered earth-fixed (ECEF) coordinate system ($e$), earth-centered inertial coordinate system ($i$), and local geodetic coordinate system ($n$). $\lambda$ and $\varphi$ are latitude and longitude of the origin of the local geodetic frame, respectively. $\boldsymbol{\omega}^e_i$ is the mean angular velocity of the Earth, and $\mathbf{g}$ is the gravity vector all expressed in $e$-frame. $\mathbf{o}_e^n$ is the vector from the origin of $e$-frame to the origin of $n$-frame. $\boldsymbol{\omega}^e_i$, $\mathbf{g}$, and $\mathbf{o}_e^n$ are expressed in $e$-frame.}
    \label{FG:ins-frame}
\end{figure}

We begin with the definition of the three coordinate systems: earth centered earth fixed (ECEF) system, local geodetic system, and vehicle body centered system. As shown in Fig.~\ref{FG:ins-frame}, earth-centered earth fixed (ECEF) system has its origin attached to the center of the Earth and rotates with it. GPS measurements are measured in the ECEF system ($e$-frame). Inertial measurements are measured in earth-centered inertial system (ECI) or $i$-frame, and are the combined result of the Earth rotation and the vehicle ego-motion. Local geodetic system ($n$-frame) has its origin coincident with the fixed ground based station, its $x$-axis always points to geodetic east, $y$-axis points to geodetic north, and $z$-axis completes the right-handed orthogonal frame. The rotation matrix from $n$-frame to $e$-frame can be written as
\begin{align}
R^e_n &= R_x(\frac{\pi}{2}-\lambda)R_z(\varphi+\frac{\pi}{2}) \nonumber\\
& = \left(
\begin{array}{ccc}
 -\text{s}_{\varphi} & -\text{c}_{\varphi} \text{s}_{\lambda} & \text{c}_{\lambda} \text{c}_{\varphi} \\
 \text{c}_{\varphi} & -\text{s}_{\lambda} \text{s}_{\varphi} & \text{c}_{\lambda} \text{s}_{\varphi} \\
 0 & \text{c}_{\lambda} & \text{s}_{\lambda}
\end{array}\right)\label{Eq:e2n_rotate}
\end{align}
For example, given a position $\mathbf{r}^e$ in $e$-frame, we write the corresponding coordinate in $n$-frame as $\mathbf{r}^n = R_e^n (\mathbf{r}^e-\mathbf{o}_e^n)$ where $R_e^n = (R^e_n)^T$ and $\mathbf{o}_e^n$ is vector from the origin of $e$-frame to the origin of $n$-frame, expressed in $e$-frame.

\begin{figure}[h]
    \centering
    \subfigure[]{\includegraphics[width=4.0cm]{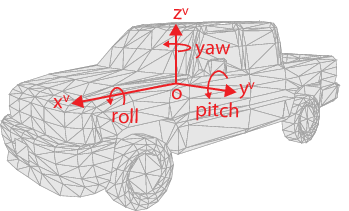}}
    \subfigure[Rotation by roll $R_x(\phi)$]{\includegraphics[width=4.0cm]{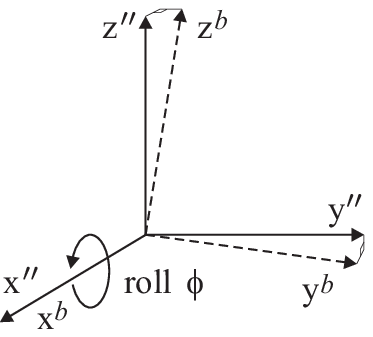}}
    \subfigure[Rotation by pitch $R_y(\theta)$]{\includegraphics[width=4.0cm]{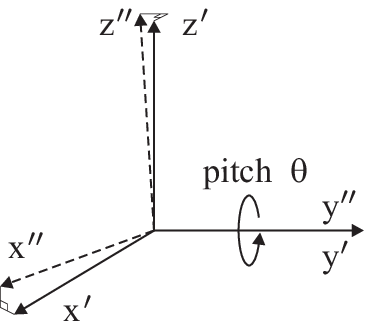}}
    \subfigure[Rotation by yaw $R_z(\psi)$]{\includegraphics[width=4.0cm]{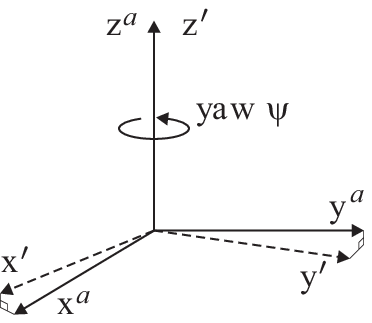}}
    \caption{\protect\small  (a) The vehicle centered coordinate frame ($v$). (b)-(d) Vehicle attitude defined by angles of roll $\phi$, pitch $\theta$, and yaw $\psi$.}
    \label{FG:yaw-pitch-roll}
\end{figure}

Fig.~\ref{FG:yaw-pitch-roll}(a) illustrates the vehicle centered system ($v$-frame). This frame has its origin at the center of gravity of the vehicle with its $x$-axis pointing in the forward direction, the $z$-axis up through the ceiling of the vehicle, and $y$-axis completes the right-handed orthogonal system.

The rotation matrix from $a$-frame (any coordinate frame $i$, $e$, $n$, or $v$) to another coordinate system $b$-frame can be derived by subsequently rotations in the three planes (see Fig.~\ref{FG:yaw-pitch-roll}(b)-(d)), i.e., first in the plane spanned by the $x$- and $y$-axis, then the one spanned by $x$- and $z$-axis, and finally the plane spanned by $y$- and $z$-axis. Mathematically, this rotation matrix can be expressed by three Euler angles $\boldsymbol{\theta}_b^a = (\psi, \theta, \phi)^T$. Note that $\boldsymbol{\theta}^b_a = -\boldsymbol{\theta}_b^a$ and
\begin{align}
R_a^b&=R_x(-\phi)R_y(-\theta)R_z(-\psi) \nonumber\\
&=\left(
\begin{array}{ccc}
 \text{c}_{\theta} \text{c}_{\psi} & \text{c}_{\theta} \text{s}_{\psi} & -\text{s}_{\theta}\\
 \text{c}_{\psi} \text{s}_{\theta} \text{s}_{\phi}-\text{c}_{\phi} \text{s}_{\psi}& \text{c}_{\phi} \text{c}_{\psi}+\text{s}_{\theta} \text{s}_{\phi} \text{s}_{\psi} & \text{c}_{\theta} \text{s}_{\phi}\\
 \text{c}_{\phi} \text{c}_{\psi}\text{s}_{\theta}+\text{s}_{\phi} \text{s}_{\psi} & -\text{c}_{\psi} \text{s}_{\phi}+\text{c}_{\phi} \text{s}_{\theta} \text{s}_{\psi} & \text{c}_{\theta} \text{c}_{\phi}
\end{array}
\right) \label{Eq:RotMatrixEulerAngle}
\end{align}
where $\text{c}_{\theta} = \cos(\theta)$ and $\text{s}_{\theta} = \sin(\theta)$.

Consider infinitesimal angles $\delta\boldsymbol{\theta}_a^b=(\delta\phi, \delta\theta, \delta\psi)^T$ for the roll, pitch, and yaw motion, the corresponding rotation matrix can be approximated by $R_a^b \approx \mathbf{I}_3 - \Delta\boldsymbol\Theta_a^b$ where $\Delta\boldsymbol{\Theta}_a^b = [\delta \boldsymbol{\theta}_a^b] \times$ is the skew symmetric matrix representation of the rotation angles $\boldsymbol\theta_a^b$, i.e.,
\[
\Delta\boldsymbol{\Theta}_a^b =\left(\begin{array}{ccc} 0 & -\delta\psi & \delta\theta \\
                              \delta\psi & 0 & -\delta\phi \\
                              -\delta\theta & \delta\phi &0
\end{array} \right)
\]

At time $t$, a vector $\mathbf{p}$ in $a$-frame can be expressed in $b$-frame as $\mathbf{q}(t) = R_a^b(t) \mathbf{p}$. Now consider at time $t+\Delta t$,
\[
\mathbf{q}(t+\Delta t) = R_a^b(t+\Delta t) \mathbf{p} = (\mathbf{I}_3 - \Delta \boldsymbol\Theta_a^b)R_a^b(t) \mathbf{p}
\]
The time derivative of the $R_a^b$ is defined as
\begin{align}
\dot{R}_a^b(t) &= \lim_{\Delta t\rightarrow 0} \frac{R_a^b(t+\Delta t) - R_a^b(t)}{\Delta t} \nonumber\\
              &= \lim_{\Delta t\rightarrow 0}  \frac{(\mathbf{I}_3 - \Delta\boldsymbol\Theta_a^b)R_a^b(t) - R_a^b(t)}{\Delta t} \nonumber\\
              &= \lim_{\Delta t\rightarrow 0} -\frac{\Delta \boldsymbol\Theta_a^b}{\Delta t} R_a^b(t) \nonumber\\
              &= -\boldsymbol{\Omega}_a^b R_a^b(t) \label{Eq:timederivative1}
\end{align}
where $\boldsymbol{\Omega}_a^b=\lim_{\Delta \rightarrow 0} \frac{\Delta \boldsymbol\Theta_a^b}{\Delta t}$ is the skew symmetric matrix angular rate $\boldsymbol\omega_a^b = (\dot\phi, \dot\theta, \dot\psi)^T$, i.e., $\boldsymbol{\Omega}_a^b=[\boldsymbol\omega_a^b]\times$.

The transpose of \eqref{Eq:timederivative1} is
\begin{equation}\dot{R}_b^a(t) = -(\boldsymbol{\Omega}_a^b R_a^b(t))^T=R_b^a(t) \boldsymbol{\Omega}_a^b \label{Eq:timederivative2}\end{equation}
where $(\boldsymbol{\Omega}_a^b)^T = - \boldsymbol{\Omega}_a^b$.

Note that \eqref{Eq:timederivative2} is equivalent to
\begin{equation}\dot{\boldsymbol{\theta}}_b^a(t) = \boldsymbol{\omega}_a^b \label{Eq:timederivative_angle}\end{equation}

\subsection{Navigation Equation}
Considering a point $\mathbf{r}^i$ in inertial frame, by Newton's laws, we can have the following kinematical acceleration equation:
\begin{equation}
\ddot{\mathbf{r}}^i = \mathbf{g}^i + \mathbf{f}^i \label{Eq:NewtonLaw}
\end{equation}
where $\mathbf{g}^i$ is the gravitational acceleration and $\mathbf{f}^i$ is the vehicle's acceleration in $i$-frame.

Assuming the center of the vehicle locate at $\mathbf{r}^n$ in the local geodetic frame, we can express the corresponding position in the inertial frame by
\begin{equation}
\mathbf{r}^i = R_e^i \mathbf{r}^e = R_e^i(R_n^e\mathbf{r}^n + \mathbf{o}^n_e) = R_n^i\mathbf{r}^n + R_e^i\mathbf{o}^n_e\label{Eq:trans_i2n}
\end{equation}
where $\mathbf{o}^n_e$ is time invariant and is the vector pointing from the origin of $e$-frame to the origin of $n$-frame, represented in $e$-frame. Note that the rotation matrix $R_n^i$ can be decomposed as $R_n^i=R_e^iR^e_n$, and $R^e_n$ is time invariant. Referring \eqref{Eq:timederivative2}, we can write derivatives of $R_n^i$ as
\begin{align}
\dot{R}_n^i  &=  R^i_n \boldsymbol{\Omega}_i^n\nonumber\\
\ddot{R}_n^i &=  R^i_n \boldsymbol{\Omega}_i^n\boldsymbol{\Omega}_i^n \nonumber
\end{align}
where
\begin{equation}
\boldsymbol{\Omega}_i^n=R_e^n\boldsymbol{\Omega}_i^e R_n^e \label{Eq:EarthRotRate-in-nframe}
\end{equation}
is the skew symmetric matrix of Earth's rotation in local geodetic frame $n$, and $\boldsymbol{\Omega}_i^e$ is the skew symmetric matrix of the Earth's angular velocity $\boldsymbol{\omega}_i^e$ defined in Fig.~\ref{FG:ins-frame}.

By differentiating \eqref{Eq:trans_i2n} twice with respect to time, we obtain
\begin{align}
\ddot{\mathbf{r}}^i &= R_n^i \ddot{\mathbf{r}}^n + 2 R_n^i \boldsymbol{\Omega}_i^n \dot{\mathbf{r}}^n + R_n^i {\boldsymbol{\Omega}}_i^n{\boldsymbol{\Omega}}_i^n (\mathbf{r}^n + \mathbf{o}^n_e) \nonumber \\
&\approx R_n^i \ddot{\mathbf{r}}^n + 2 R_n^i \boldsymbol{\Omega}_i^n \dot{\mathbf{r}}^n +R_n^i {\boldsymbol{\Omega}}_i^n{\boldsymbol{\Omega}}_i^n \mathbf{o}^n_e \label{eq:naveq_ddot_r}
\end{align}
where we assume $\|\mathbf{o}^n_e\|\gg \|\mathbf{r}^n\|$.

Plugging \eqref{Eq:NewtonLaw} into \eqref{eq:naveq_ddot_r} and multiplying $R_i^n$ to both sides, we can approximate \eqref{Eq:NewtonLaw} to be
\begin{equation}
 \mathbf{g}^n + \mathbf{f}^n = \ddot{\mathbf{r}}^n + 2  \boldsymbol{\Omega}_i^n \dot{\mathbf{r}}^n + ({\boldsymbol{\Omega}}_i^n)^2 \mathbf{o}^n_e
\label{eq:naveq_nframe}
\end{equation}
where $\mathbf{f}^n$ is the vehicle's acceleration in $n$-frame, and $\mathbf{g}^n$ is the gravity vector, $\mathbf{g}^n = (0, 0, -9.80665)^T$ $\text{m}/\text{s}^2$ \cite{UsefulConstants}.

Now we consider the kinematics of the vehicle attitude, expressed as the rotation matrix $R_v^n$ from $v$-frame to $n$-frame.
\begin{align}
\dot{R}_v^n &= R_v^n \boldsymbol{\Omega}_n^v \label{eq:naveq_attitude}
\end{align}
where the skew-symmetric matrix $\boldsymbol{\Omega}_n^v$ for the the rotation rates between the local geodetic and vehicle frames consists of the angular rates $\boldsymbol{\omega}_i^v$ measured by the gyros and the Earth rotational rate in $n$-frame, i.e., $\boldsymbol{\omega}_n^v =\boldsymbol{\omega}_i^v - R_e^v \boldsymbol{\omega}_i^e$.

Note that the rotation matrix  $R_v^n$ can be expressed by roll-pitch-yaw angles $\boldsymbol{\theta}_v^n = (\phi, \theta, \psi)^T$ (c.f., \eqref{Eq:RotMatrixEulerAngle}). Referred to \eqref{Eq:timederivative_angle}, we note that \eqref{eq:naveq_attitude} is equivalent to
\begin{align}
\dot{\boldsymbol{\theta}}_v^n &= \boldsymbol{\omega}_i^v - R_e^v \boldsymbol{\omega}_i^e  \label{eq:naveq_attitudeEulerAngle}
\end{align}
where $\boldsymbol{\omega}_i^e=(0,0,7.29211501\times 10^{-5})$ rad/s \cite{UsefulConstants} is the mean angular velocity of the Earth (c.f., Fig.~\ref{FG:ins-frame}).

In summary, combining Eqs.\eqref{eq:naveq_nframe} and \eqref{eq:naveq_attitudeEulerAngle}, we obtain the navigation equation in the first-order differential equations as
\begin{equation}
\left[\begin{array}{c}
\dot{\mathbf{r}}^n\\
\dot{\mathbf{v}}^n\\
\dot{\boldsymbol{\theta}}_v^n
\end{array}\right] = \left[ \begin{array}{c}
\mathbf{v}^n\\
-2 \boldsymbol{\Omega}_i^n \mathbf{v}^n - ({\boldsymbol{\Omega}}_i^n)^2\mathbf{o}^n_e  + \mathbf{g}^n + R_v^n \mathbf{f}^v\\
\boldsymbol{\omega}_i^v - R_e^v \boldsymbol{\omega}_i^e
\end{array}
\right] \label{Eq:naveq}
\end{equation}
where $\mathbf{v}^n$ is the vehicle velocity in $n$-frame; and $\mathbf{f}^v$ and $\boldsymbol{\omega}_i^v$ are vehicle acceleration and angular rate in $v$-frame, which is directly measured by the accelerometers and gyros, respectively.

Note that $-2 \boldsymbol{\Omega}_i^n \mathbf{v}^n$ and $-({\boldsymbol{\Omega}}_i^n)^2\mathbf{o}^n_e$ in \eqref{Eq:naveq} are the Coriolis and centrifugal terms induced by the rotation of the Earth, and $\boldsymbol{\Omega}_i^n$ is defined in \eqref{Eq:EarthRotRate-in-nframe}.

\section{GPS Data Processing} \label{SC:GPSprocessing}

In this section, we develop the processing necessary to use GPS measurements for relative positioning. Consider the reference point $A$ (base station) and the rover point $B$ (center of the receiving antenna in the vehicle) in $e$-frame (c.f., Section~\ref{SC:IMUprocessing}). Let $\mathbf{r}_A =[X_A, Y_A, Z_A]^T$ and $\dot{\mathbf{r}}_A=[\dot X_A, \dot Y_A, \dot Z_A]^T$ denote the position and velocity vectors of $A$, respectively; $\mathbf{r}_B = [X_B, Y_B, Z_B]^T$ and $\dot{\mathbf{r}}_B = [\dot X_B, \dot Y_B, \dot Z_B]^T$ denote the position and velocity vectors of $B$, respectively. The baseline vector can be written as
\[
\begin{array}{ll}
\mathbf{b} = \mathbf{r}_B - \mathbf{r}_A,  &
\dot{\mathbf{b}} = \dot{\mathbf{r}}_B - \dot{\mathbf{r}}_A
\end{array}
\]
Note that in this paper we use the reference point $A$ as the origin of the $n$-frame. Namely, $\mathbf{r}_A = \mathbf{o}^n_e$ is the origin of the $n$-frame in the $e$-frame.

\subsection{GPS Observations}
The three basic measurements of a GPS receiver from a satellite are code (pseudo-range), phase, and doppler. For short baseline relative positioning\footnote{This refers to a relative distance between base and vehicle of 10 km for single frequency or 50 km for dual frequency under most atmospheric conditions~\cite{Strang1997}.} the accuracy could be substantially improved by having a receiver (reference) broadcast its measurements to nearby receivers (rovers). Let $*^{(\alpha)}_{\beta}$ denote the measurement from the receiver $\beta$ and the $\alpha$-th satellite. Giving two receivers $A$ (reference) and $B$, and two satellites $j$ (reference), and $k$, we define the double-difference convention $*^{(jk)}_{AB} = *^{(k)}_{B} - *^{(k)}_{A} - *^{(j)}_{B} + *^{(j)}_{A}$ where the asterisk may be replaced by $R$, $\Phi$, $D$, $\rho$, and $\dot \rho$ that correspond to pseudo-range measurement, phase measurement, Doppler measurement, geometric distance between receiver and satellite, and time rate of the geometric distance. Thus the double-difference measurements (c.f., \cite[p. 460]{Strang1997}) can be written as
\begin{eqnarray}
R_{AB}^{(jk)} &=& \rho_{AB}^{(jk)} + \eta_{AB}^{(k)} - \eta_{AB}^{(j)} \label{Eq:L1_codedouble}\\
\lambda \Phi_{AB}^{(jk)} &=& \rho_{AB}^{(jk)}  + \lambda (a_{AB}^{(k)} - a_{AB}^{(j)}) + \xi_{AB}^{(k)}-\xi_{AB}^{(j)} \label{Eq:L1_phasedouble}\\
-\frac{c D_{AB}^{(jk)}}{f} &=& \dot{\rho}_{AB}^{(jk)} + \zeta_{AB}^{(k)}-\zeta_{AB}^{(j)} \label{Eq:L1_dopplerdouble}
\end{eqnarray}
where the symbols $R_{AB}^{(jk)}$, $\Phi_{AB}^{(jk)}$ and $D_{AB}^{(jk)}$ denote the double-differences of code, phase, doppler measurements between the rover receiver $B$ and base receiver $A$, respectively;  $\rho_{AB}^{(jk)}$  is the double-difference geometric distance $\rho_{AB}^{(jk)}=\rho_B^{(k)} - \rho_A^{(k)}-\rho_B^{(j)}+\rho_B^{(j)}$; $\dot\rho_{AB}^{(jk)}$ is the time derivatives of $\rho_{AB}^{(jk)}$; single-difference $a_{AB}^{(\alpha)}$ is the ambiguity for the $\alpha$-th satellite\footnote{ $a_{AB}^{(\alpha)}$ at time step $t$ corresponds to the $\kappa_{\alpha,t}$-th component in the ambiguity vector $\mathbf{a}$.}; $\lambda$ and $f$ are the carrier wavelength and frequency, respectively; $c$ is the speed of light; single-differences $\eta_{AB}^{(\alpha)}$, $\xi_{AB}^{(\alpha)}$, and $\zeta_{AB}^{(\alpha)}$ are the corresponding measurement errors.

We assume $\eta_{AB}^{(\alpha)}$, $\xi_{AB}^{(\alpha)}$, and $\zeta_{AB}^{(\alpha)}$ are unbiased and independently distributed with Gaussian distribution for different satellites at different epochs (c.f., \eqref{Eq:pseudorange_errmodel}-\eqref{Eq:doppler_errmodel}).

In \eqref{Eq:L1_codedouble}-\eqref{Eq:L1_dopplerdouble}, we consider only the single carrier frequency ($f_1=1575.42$MHz, $\lambda_1 = c/f_1$) since most low-cost receivers only receive L1 signals. The case of dual-frequency may easily be accommodated by adding three more measurements as \eqref{Eq:L1_codedouble}-\eqref{Eq:L1_dopplerdouble} with $f_2=1227.60$MHz and $\lambda_2 = c/f_2$, resulting in six basic outputs and two ambiguities per satellite. Also in the similar fashion we can handle the wide lane combination.

\subsection{Relative Positioning}
Let $\mathbf{b}_0$ denote the approximated baseline. Let $\mathbf{r}^{(j)} = [X^{(j)}, Y^{(j)}, Z^{(j)}]^T$ and $\mathbf{r}^{(k)} = [X^{(k)}, Y^{(k)}, Z^{(k)}]^T$ denote the earth-rotation corrected positions of the $j$-th and $k$-th satellites in the ECEF frame, respectively. Let $X_{B_0}$, $Y_{B_0}$, and $Z_{B_0}$ be the component values of the approximated position $\mathbf{r}_{B_0}$ ($\mathbf{r}_{B_0} = \mathbf{r}_A + \mathbf{b}_0$) for the unknown point $B$. Then, the approximated geometric distances between the point $B$ and the satellites $j$ and $k$ can be calculated as
{\small\begin{equation}
\begin{split}
\rho_{B_0}^{(j)} &= \sqrt{ (X^{(j)}-X_{B_0})^2 + (Y^{(j)}-Y_{B_0})^2 + (Z^{(j)}-Z_{B_0})^2} \\
\rho_{B_0}^{(k)} &= \sqrt{ (X^{(k)}-X_{B_0})^2 + (Y^{(k)}-Y_{B_0})^2 + (Z^{(k)}-Z_{B_0})^2}
\end{split}
\nonumber\end{equation}}The distances between the point $A$ and satellites $j$ and $k$ can be calculated as
{\small\begin{equation}
\begin{split}
\rho_{A}^{(j)} &= \sqrt{ (X^{(j)}-X_{A})^2 + (Y^{(j)}-Y_{A})^2 + (Z^{(j)}-Z_{A})^2} \\
\rho_{A}^{(k)} &= \sqrt{ (X^{(k)}-X_{A})^2 + (Y^{(k)}-Y_{A})^2 + (Z^{(k)}-Z_{A})^2}
\end{split}
\nonumber
\end{equation}}

We define $\dot\rho_{\beta}^{(i)}$ and $\mathbf{n}_{\beta}^{(i)}$ as the range rate and the vector of unit length from the receiver $\beta$ ($\beta=A$ or $B$) to the $i$-th satellite, respectively. Let $\dot{\mathbf{b}}_0$ be the approximated baseline velocity. Let $\dot{\mathbf{r}}^{(j)}$ and $\dot{\mathbf{r}}^{(k)}$ denote velocity vectors of satellites $j$ and $k$, respectively, and $\dot{\mathbf{r}}_{B_0}$ ($\dot{\mathbf{r}}_{B_0} = \dot{\mathbf{r}}_{A} + \dot{\mathbf{b}}_0$) the approximated velocity for the unknown point $B$. Then, the range rates $\dot \rho_{B_0}^{(j)}$, $\dot \rho_{B_0}^{(k)}$, $\dot \rho_{A}^{(j)}$, and $\dot \rho_{A}^{(k)}$ are computed as
{\small\begin{equation}
\begin{split}
\dot \rho_{B_0}^{(j)} &= (\dot{\mathbf{r}}_{B_0} - \dot{\mathbf{r}}^{(j)})^T \mathbf{n}_{B_0}^{(j)} \\
\dot \rho_{B_0}^{(k)} &= (\dot{\mathbf{r}}_{B_0} - \dot{\mathbf{r}}^{(k)})^T \mathbf{n}_{B_0}^{(k)} \\
\dot \rho_{A}^{(j)} &= (\dot{\mathbf{r}}_{A} - \dot{\mathbf{r}}^{(j)})^T \mathbf{n}_{A}^{(j)} \\
\dot \rho_{A}^{(k)} &= (\dot{\mathbf{r}}_{A} - \dot{\mathbf{r}}^{(k)})^T \mathbf{n}_{A}^{(k)} \\
\end{split}
\nonumber
\end{equation}}

When the models in \eqref{Eq:L1_codedouble}-\eqref{Eq:L1_dopplerdouble} are considered, the only terms comprising unknowns in nonlinear form are $\rho_{AB}^{(jk)}$ and $\dot{\rho}_{AB}^{(jk)}$. Here we outline how $\rho_{AB}^{(jk)}$ and $\dot{\rho}_{AB}^{(jk)}$ is linearized in term of $\mathbf{b}$ and $\dot{\mathbf{b}}$.

Since $\mathbf{r}_A$ and $\dot{\mathbf{r}}_A$ of the reference point $A$ are known (using single point solution for the reference receiver), we can write the linearized $\rho_{AB}^{(jk)}$ in the neighborhood of $\mathbf{b}_0$ using Taylor expansion as
\begin{equation}
\begin{array}{ll}
\rho_{AB}^{(jk)} &=\rho_{B}^{(k)} - \rho_{A}^{(k)}  - \rho_{B}^{(j)} + \rho_{A}^{(j)} \\
 &=   \rho_{AB_0}^{(jk)} + (\Lambda_{AB_0}^{(jk)})^T  (\mathbf{b}-\mathbf{b}_0) + \mbox{h.o.t.}
\end{array}
\label{Eq:DD_geometry_range}
\end{equation}
where \[\Lambda_{AB_0}^{(jk)} = \left[
  \begin{array}{c} -\frac{X^{(k)}-X_{B_0}}{\rho_{B_0}^{(k)}} + \frac{X^{(j)}-X_{B_0}}{\rho_{B_0}^{(j)}} \\
  -\frac{Y^{(k)}-Y_{B_0}}{\rho_{B_0}^{(k)}} + \frac{Y^{(j)}-Y_{B_0}}{\rho_{B_0}^{(j)}}\\
  - \frac{Z^{(k)}-Z_{B_0}}{\rho_{B_0}^{(k)}} + \frac{Z^{(j)}-Z_{B_0}}{\rho_{B_0}^{(j)}}
  \end{array}\right]
\]\[\rho_{AB_0}^{(jk)} = \rho_{B_0}^{(k)}- \rho_{A}^{(k)} - \rho_{B_0}^{(j)} + \rho_{A}^{(j)}\] and h.o.t. represents the higher order terms of Taylor expansion.

Similarly, we can write $\dot{\rho}_{AB}^{(jk)}$ as
\begin{equation}
\begin{array}{ll}
\dot\rho_{AB}^{(jk)} &= \dot\rho_{B}^{(k)} -\dot\rho_{A}^{(k)} - \dot\rho_{B}^{(j)} + \dot\rho_{A}^{(j)} \\
 &= \dot \rho_{AB_0}^{(jk)} + l^{(jk)}_{AB_0}+ (\beta_{AB_0}^{(jk)})^T (\dot{\mathbf{b}}-\dot{\mathbf{b}}_0)
\end{array}
\label{Eq:DD_geometry_doppler}
\end{equation}
where
\[\beta_{AB_0}^{(jk)} = -\left[
  \begin{array}{c} \frac{X^{(k)}-X_{B_0}}{\rho_{B_0}^{(k)}}  - \frac{X^{(j)}-X_{B_0}}{\rho_{B_0}^{(j)}} \\
  \frac{Y^{(k)}-Y_{B_0}}{\rho_{B_0}^{(k)}} -   \frac{Y^{(j)}-Y_{B_0}}{\rho_{B_0}^{(j)}} \\
   \frac{Z^{(k)}-Z_{B_0}}{\rho_{B_0}^{(k)}} - \frac{Z^{(j)}-Z_{B_0}}{\rho_{B_0}^{(j)}}
  \end{array}\right] \]\[\dot\rho_{AB_0}^{(jk)} = \dot\rho_{B_0}^{(k)}- \dot\rho_{A}^{(k)} - \dot\rho_{B_0}^{(j)} + \dot\rho_{A}^{(j)}\]{\small \[
\begin{split}
&l^{(jk)}_{AB_0} = \\
&\begin{array}{l} \left(-\frac{X^{(k)} - X_{B_0}}{\rho^{(k)}_{B_0}}+\frac{X^{(k)} - X_{A}}{\rho^{(k)}_{A}} + \frac{X^{(j)} - X_{B_0}}{\rho^{(j)}_{B_0}}-\frac{X^{(j)} - X_{A}}{\rho^{(j)}_{A}}\right)\dot X_A \\
+\left(-\frac{Y^{(k)} - Y_{B_0}}{\rho^{(k)}_{B_0}}+\frac{Y^{(k)} - Y_{A}}{\rho^{(k)}_{A}} + \frac{Y^{(j)} - Y_{B_0}}{\rho^{(j)}_{B_0}}-\frac{Y^{(j)} - Y_{A}}{\rho^{(j)}_{A}}\right)\dot Y_A \\
+\left(-\frac{Z^{(k)} - Z_{B_0}}{\rho^{(k)}_{B_0}}+\frac{Z^{(k)} - Z_{A}}{\rho^{(k)}_{A}} + \frac{Z^{(j)} - Z_{B_0}}{\rho^{(j)}_{B_0}}-\frac{Z^{(j)} - Z_{A}}{\rho^{(j)}_{A}}\right)\dot Z_A
\end{array}
\end{split}\]}

\subsection{Measurement Matrix} {\label{SC:measurement_matrix}
Suppose there are $M$ visible satellites. Without loss of generality we choose satellite 1 as the reference satellite (i.e., $j=1$).  We define $J= \left[\begin{array}{cc}-\mathbf{e} & \mathbf{I}_{M-1}\end{array}\right] =\left[\begin{array}{ccc} J_2^T&\cdots& J_M^T\end{array} \right]^T $ with $\mathbf{e}=[\underbrace{1,...,1}_{M-1}]^T$ and $\mathbf{I}_{M-1}$ an identity matrix.
One can verify that $J$ is a $(M-1) \times M$ matrix and its columns are linearly dependent. We also define
\begin{eqnarray}
\mathbf{y} &=& \left[\begin{array}{cc}\mathbf{b} & \dot{\mathbf{b}}\end{array}\right]^T\nonumber\\
\mathbf{a} &=& \left[\begin{array}{ccc}a_{AB}^{1}&...&a_{AB}^{M}\end{array}\right]^T\nonumber
\end{eqnarray}
\begin{equation}
\begin{array}{ll}
E_R = \left[\begin{array}{cc}(\Lambda_{AB_0}^{(12)})^T& \mathbf{0}_{1\times 3} \\ \vdots & \vdots \\ (\Lambda_{AB_0}^{(1M)})^T & \mathbf{0}_{1\times 3} \end{array}\right] &
E_D = \left[\begin{array}{cc} \mathbf{0}_{1\times 3} & (\beta_{AB_0}^{(12)})^T\\ \vdots & \vdots \\ \mathbf{0}_{1\times 3} & (\beta_{AB_0}^{(1M)})^T \end{array}\right]
\end{array}\nonumber
\end{equation}
\begin{eqnarray}
\mathbf{o}_R &=& \left[\begin{array}{c} R_{AB}^{(12)}-\rho_{AB_0}^{(12)}+  (\Lambda_{AB_0}^{(12)})^T \mathbf{b}_0 \\ \vdots \\ R_{AB}^{(1M)}-\rho_{AB_0}^{(1M)} + (\Lambda_{AB_0}^{(1M)})^T \mathbf{b}_0\end{array}\right]\nonumber\\
\mathbf{o}_\Phi &=& \left[\begin{array}{c} \lambda \Phi_{AB}^{(12)}-\rho_{AB_0}^{(12)}+ (\Lambda_{AB_0}^{(12)})^T \mathbf{b}_0   \\ \vdots \\ \lambda \Phi_{AB}^{(1M)}-\rho_{AB_0}^{(1M)} + (\Lambda_{AB_0}^{(1M)})^T\mathbf{b}_0  \end{array}\right]\nonumber\\
\mathbf{o}_D &=& \left[\begin{array}{c} -\frac{c D_{AB}^{(12)}}{f}- \dot\rho_{AB_0}^{(12)}-l^{(12)}_{AB_0}+(\beta_{AB_0}^{(12)})^T\mathbf{b}_0\\ \vdots \\-\frac{c D_{AB}^{(1M)}}{f}- \dot\rho_{AB_0}^{(1M)} - l^{(1M)}_{AB_0}+(\beta_{AB_0}^{(1M)})^T\mathbf{b}_0\end{array}\right]\nonumber
\end{eqnarray}
Ignoring the h.o.t. terms in \eqref{Eq:DD_geometry_range} and plugging \eqref{Eq:DD_geometry_range} and \eqref{Eq:DD_geometry_doppler} into \eqref{Eq:L1_codedouble}-\eqref{Eq:L1_dopplerdouble}, we combine $M-1$ double-differences for code, phase, and doppler measurements as
\begin{eqnarray}
\mathbf{o}_R &=& E_R \mathbf{y} + \nu_\eta \label{Eq:DD_code_correl}\\
\mathbf{o}_\Phi &=& E_R \mathbf{y} + \lambda J \mathbf{a} + \nu_\xi  \label{Eq:DD_phase_correl}\\
\mathbf{o}_D &=& E_D \mathbf{y} + \nu_\zeta \label{Eq:DD_doppler_correl}
\end{eqnarray}
where $\nu_\eta$, $\nu_\xi$, and $\nu_\zeta$ are the corresponding noise vectors for code, phase, and doppler measurements, respectively.

Note that the noise vectors $\nu_\eta$, $\nu_\xi$, and $\nu_\zeta$ are correlated.
Let $L_J$ be the Cholesky factor of $P_J^{-1}$, i.e., $P_J^{-1}=L_J^TL_J$ where \[
P_J = \left[\begin{array}{cccc} 2 & 1 & \cdots & 1 \\
1 & 2 & \cdots & 1 \\
\vdots & \vdots & \ddots & \vdots \\
1 & 1 & \cdots & 2
\end{array}\right]
\]
is a $M-1 \times M-1$ matrix.
One can verify that in information array form\footnote{The information array is an alternative representation of the Gaussian distribution. Rather than using the mean and covariance as the parameters of a Gaussian distribution (i.e., $\mathcal{N}(x; \mu, \Sigma)$), we instead parameterize in the square root of the information matrix $\mathcal{I} = \Sigma^{-1} = R R^T$ and the normalized information vector $\mathbf{z} = R \mu$, i.e., $p(x) \sim [R, \mathbf{z}]$.} the distribution of $v_\eta$, $v_\xi$, and $v_\zeta$ can be written as
\begin{equation}
\begin{array}{lll}
\nu_\eta \sim [ L_J U_R, \mathbf{0}], &
\nu_\xi \sim [L_J U_\Phi, \mathbf{0}], &
\nu_\zeta \sim [L_J U_D, \mathbf{0}]\nonumber
\end{array}
\end{equation}
where $U_R = \text{diag}[u_R^1,...,u_R^M]$, $U_R = \text{diag}[u_\Phi^1,...,u_\Phi^M]$, $U_D = \text{diag}[u_D^1,...,u_D^M]$, and $u_R^\alpha$, $u_\Phi^\alpha$, and $u_D^\alpha$ are defined in \eqref{Eq:pseudorange_errmodel}-\eqref{Eq:doppler_errmodel}, respectively.

\eqref{Eq:DD_code_correl}-\eqref{Eq:DD_doppler_correl} can be de-correlated by multiplying $L_J U_R$, $L_J U_\Phi$, and $L_J U_D$ on both sides, respectively
\begin{align}
L_J U_R \mathbf{o}_R &= L_J U_R E_R \mathbf{y} + \tilde\nu_\eta \label{Eq:DD_code_decorrel}\\
L_J U_\Phi \mathbf{o}_\Phi &= L_J U_\Phi E_\Phi \mathbf{y} + \lambda L_J U_\Phi J \mathbf{a} + \tilde\nu_\xi  \label{Eq:DD_phase_decorrel}\\
L_J U_D \mathbf{o}_D &= L_J U_D E_D  \mathbf{y} + \tilde\nu_\zeta \label{Eq:DD_doppler_decorrel}
\end{align}
where $\tilde\nu_\eta \sim \mathcal{N}(\mathbf{0},\mathbf{I}_{M-1})$, $\tilde\nu_\xi \sim \mathcal{N}(\mathbf{0},\mathbf{I}_{M-1})$, and $\tilde\nu_\zeta \sim \mathcal{N}(\mathbf{0},\mathbf{I}_{M-1})$.

\section{Sensor Error Models} \label{SC:SensorErrorModel}
In this section we model the errors generated by IMU, speedometer, and GPS measurements using a stochastic model. Errors of IMU and vehicle velocity measurements are modeled by first-order Gaussian Markov stochastic process. The errors of pseudo-range measurements are zero-mean Gaussian distribution with the variance being a function of signal-to-noise ratio (SNR). The errors of phase and Doppler measurements are zero-mean Gaussian distribution, and the variance depends on satellite's elevation angle.

\subsection{IMU Sensor Errors}
All measurement from sensors is degraded because of errors. The primary sources of errors for IMU sensors are bias, scale factor, and measurement noise. Some errors are contributed from deterministic process and can be corrected through specific bench-calibration procedures, while the other errors are not deterministic and need to be modeled by a stochastic process. The accelerometer or gyro measurement can be expressed as
\begin{equation}
s(t) = (1+S)\tilde{s}(t)  + \beta(t) \nonumber
\end{equation}
where $s$ and $\tilde{s}$ are the true value of the quantity to be measured and the sensor's measured output, respectively; $S$ is the scale factor error; and $\beta(t)$ is the bias.

The bias term $\beta(t)$ can be decomposed as the following two terms:
\[
\beta(t) = \beta_0 + \beta_1(t)
\]
where $\beta_0$ represents the time-invariant component, and $\beta_1(t)$ represents the time varying component. $\beta_0$ is usually specified on IMU sensor data sheets as the ``turn-on to turn-off'' bias variation.

The time varying component $\beta_1(t)$ is typically model as a first-order Gaussian Markov stochastic process~\cite{Demoz2004,Flenniken2005}, which is expressed as the following ordinary differential equation:
\begin{equation}
\dot{\beta}_1 = -\frac{1}{\tau} \beta_1  + \sigma_\beta w \label{Eq:SDE}
\end{equation}
where $\sigma_\beta$ is the standard deviation of {\em random walk} specified in the sensor's data sheet, and $w$ is a Gaussian distribution, i.e., $w \sim \mathcal{N}(0,1)$.

The discrete version of \eqref{Eq:SDE} can be expressed as
\begin{equation}
\beta_1 (t+1) = (1-\frac{\Delta t}{\tau}) \beta_1(t) + \sigma_\beta\sqrt{\Delta t} w
\label{Eq:SDE_discrete}
\end{equation}
where $\beta_1(t+1)$ and $\beta_1(t)$ are the bias at time step $t+1$ and $t$, respectively, and $\Delta t$ is time interval between two contiguous steps.

For most survey-grade IMU sensors, we note that the time constant $\tau \gg \Delta t$, and the term in \eqref{Eq:SDE_discrete} related to $\tau$ can be neglected. Adding $\beta_0$ to the both sides of \eqref{Eq:SDE_discrete}, we have the process equation for sensor bias as
\begin{equation}
\beta(t+1) = \beta(t) + \sigma_\beta\sqrt{\Delta t} w \label{Eq:bias_dynamics}
\end{equation}

Similarly, we have the process equation for scale factor as
\begin{equation}
S(t+1) = S(t) + \sigma_S\sqrt{\Delta t} w \label{Eq:scale_dynamics}
\end{equation}
where $\sigma_S$ the sum of standard deviation of {\em scale factor error} in the sensor's data sheet

\subsection{Speedometer Measurement Error} \label{sc:wheelencoderErr}
The speedometer measurement of the land vehicle can be measured by wheel encoders.
\begin{equation}
\tilde{v}_H = v_H + S_v v_H  +  \beta_v
\label{Eq:encoder_errmodel}
\end{equation}
where $v_H$ and $\tilde{v}_H$ are the true ground and measured vehicular ground velocity, respectively;  $\beta_v$ is the {\em random walk} term modeling the measurement bias; and $S_v$ is the scale factor of the velocity measurement.

\subsection{GPS Measurement Errors}
The three GPS measurements of pseudo-range, phase, and doppler from a satellite can be modeled as \eqref{Eq:L1_codedouble}-\eqref{Eq:L1_dopplerdouble} where $\eta_{AB}^{(\alpha)}$, $\xi_{AB}^{(\alpha)}$, and $\zeta_{AB}^{(\alpha)}$ are the error terms, respectively.

\subsubsection{Pseudorange error $\eta_{AB}^{(\alpha)}$}
The single difference of pseudorange error for the $\alpha$-th satellite, $\eta_{AB}^{(\alpha)}$ can be expressed as
\[
\eta_{AB}^{(\alpha)} = \eta_{B}^{(\alpha)} - \eta_{A}^{(\alpha)}
\]
Using the variance model of \cite{AWieser2005}, we model $\eta_{\beta}^{(\alpha)}$ ($\beta=A$ or $B$) as a zero-mean Gaussian distribution $\eta_{\beta}^{(\alpha)} \sim \mathcal{N}(0, C \cdot 10^{-\frac{\text{SNR}_\beta^{(\alpha)}}{10}})$ where the variance is a function of signal-to-noise ratio $\text{SNR}_\beta$ and $C = 0.7 \cdot 10^5$ $\text{m}^2$.
Assuming $\eta_{A}^{(\alpha)}$ and $ \eta_{A}^{(\alpha)}$ are independent, we can write the variance of $\eta_{AB}^{(\alpha)}$ as $C \cdot (10^{-\frac{\text{SNR}_A^{(\alpha)}}{10}} + 10^{-\frac{\text{SNR}_B^{(\alpha)}}{10}})$, and the distribution of $\eta_{AB}^{(\alpha)}$ in information array form can be written as
\begin{equation}
\eta_{AB}^{(\alpha)} \sim [u_R^\alpha, \mathbf{0}] \label{Eq:pseudorange_errmodel}
\end{equation}
where $u_R^\alpha = \frac{1}{\sqrt{C \cdot (10^{-\frac{\text{SNR}_A^{(\alpha)}}{10}} + 10^{-\frac{\text{SNR}_B^{(\alpha)}}{10}})}}$.

\subsubsection{Phase error $\xi_{AB}^{(\alpha)}$}
We model the phase error $\xi_{AB}^{(\alpha)} \sim \mathcal{N}(0, \frac{\sigma_\Phi^2}{\sin^2 E_\alpha})$ where $E_\alpha$ is the elevation angle of the satellite $\alpha$, and $\sigma_\Phi^2$ is the measurement variance. In information array form, the distribution is
\begin{equation}
\xi_{AB}^{(\alpha)} \sim [u_\Phi^\alpha, \mathbf{0}] \label{Eq:phase_errmodel}
\end{equation}
where $u_\Phi^\alpha = \frac{\sin E_\alpha}{\sigma_\Phi}$.

\subsubsection{Doppler error $\zeta_{AB}^{(\alpha)}$}
We model the Doppler measurement error $\zeta_{AB}^{(\alpha)} \sim \mathcal{N}(0, \frac{\sigma_D^2}{\sin^2 E_\alpha})$ with $\sigma_D^2$ being the variance for doppler measurement. In information array form, the distribution is
\begin{equation}
\zeta_{AB}^{(\alpha)} \sim [u_D^\alpha, \mathbf{0}] \label{Eq:doppler_errmodel}
\end{equation}
where $u_D^\alpha =\frac{\sin E_\alpha}{\sigma_D}$.

\section{IMU and GPS Integration} \label{SC:IMU_GPS_Integration}
In this section, we provide the implementation details how to integrate data from IMU, GPS, and speedometer using BN \cite{Zeng_TIM2009}.

\subsection{GPS Measurement Equation}
We have expressed the unknown state vector $\mathbf{y}$ in $e$-frame in Section~\ref{SC:GPSprocessing}. However, the local geodetic coordinate system ($n$-frame in Section~\ref{SC:IMUprocessing}) with the reference receiver $\mathbf{r}_A$ as the origin is more appropriate to integrate with data from IMU and in-vehicle sensor.

Let $\mathbf{r}_{\text{GPS}} = R_e^n \mathbf{b}$ and $\mathbf{v}_{\text{GPS}} = R_e^n \dot{\mathbf{b}}$ where the rotation matrix $R_e^n$ is defined in \eqref{Eq:e2n_rotate}.

Usually IMU center and GPS antenna are not placed at the same position on the vehicle. This spatial separation causes the IMU and GPS measurements to be slightly different in position and velocities. This effect is called {\em level-arm} effect and can be modeled as the following equation position
\begin{align}
\mathbf{r}_{\text{GPS}} = \mathbf{r}^n + R_v^n \Delta\mathbf{r}^v_{\text{LA}} \label{Eq:levelarm_position}
\end{align}
where $\mathbf{r}^n$ is the position of the IMU center in \eqref{Eq:naveq}.

Taking derivative of \eqref{Eq:levelarm_position} with respect to time, we obtain
\begin{align}
\mathbf{v}_{\text{GPS}} &= \mathbf{v}^n + \dot{R}_v^n \Delta\mathbf{r}^v_{\text{LA}} \nonumber
\end{align}
where $\dot{R}_v^n$ is computed in \eqref{eq:naveq_attitude} and $\mathbf{v}^n$ is the velocities of the IMU center in \eqref{Eq:naveq}.

In this paper we assume $\Delta\mathbf{r}^v_{\text{LA}}$ is known by surveying.

We define the augmented state vector $\mathbf{x}$ consisting of kinematic vector of the IMU center in $n$-frame and terms for compensating bias and scale factor, i.e.,
\begin{equation}
\mathbf{x} = \left[\begin{array}{ccccccc}\mathbf{r}^n&\mathbf{v}^n & \boldsymbol{\theta}_v^n & \boldsymbol{\beta}_f & \boldsymbol{\beta}_\omega & \mathbf{S}_f & \mathbf{S}_\omega \end{array}\right]^T
\nonumber
\end{equation}
where $\boldsymbol{\theta}_v^n$ is the vehicle's attitude with respect to $n$-frame, $\boldsymbol{\beta}_f = (b_{f_x}, b_{f_y}, b_{f_z})^T$ are the bias terms for the accelerometers along $x$-, $y$-, and $z$-axis and $\boldsymbol{\beta}_\omega=(b_{\omega_{\phi}}, b_{\omega_{\theta}}, b_{\omega_{\psi}})^T$ for gyros rates along $x$-, $y$-, and $z$-axis in the vehicle frame, respectively; and $\mathbf{S}_f= (S_{f_x}, S_{f_y}, S_{f_z})^T$ and $\mathbf{S}_\omega=(S_{\omega_{\phi}}, S_{\omega_{\theta}},S_{\omega_{\psi}})^T$ are the scale factors for the corresponding measurements by the IMU sensors.

Therefore we can express GPS antenna state vector $\mathbf{y}$ in $e$-frame by the augmented state $\mathbf{x}$
\begin{align}
\mathbf{y} = \Gamma \mathbf{x} \nonumber
\end{align}
where $\Gamma = \left[\begin{array}{ccccccc} R_n^e & \mathbf{0}_3 & \mathbf{0}_3 & \mathbf{0}_3 & \mathbf{0}_3 & \mathbf{0}_3 & \mathbf{0}_3\\
             \mathbf{0}_3 & R_n^e & \mathbf{0}_3 & \mathbf{0}_3 & \mathbf{0}_3 & \mathbf{0}_3 & \mathbf{0}_3
      \end{array}\right]$
and $\mathbf{0}_3$ is a zero-valued matrix with size of $3 \times 3$.

Let
\begin{equation}
\mathbf{o} = \left[ \begin{array}{c} L_J U_R \mathbf{o}_R \\
L_J U_\Phi \mathbf{o}_\Phi \\
L_J U_D \mathbf{o}_D
\end{array}\right]  -
\left[\begin{array}{c} L_J U_R E_R \\
L_J U_\Phi E_R \\
L_J U_D E_D
\end{array}\right] \left[\begin{array}{c} R_v^e \\ R_n^e \dot{R}_v^n \end{array}\right]\Delta \mathbf{r}_{\text{LA}} \nonumber\\
\end{equation}
and
\begin{equation}
\begin{array}{ll}
H_x = \left[\begin{array}{c} L_J U_R E_R \\
L_J U_\Phi E_R \\
L_J U_D E_D
\end{array}\right]\Gamma,
&
H_a = \left[\begin{array}{c} 0\\
\lambda L_J U_\Phi J \\
0
\end{array}\right]
\end{array} \nonumber
\end{equation}
We can write measurement equations \eqref{Eq:DD_code_decorrel}-\eqref{Eq:DD_doppler_decorrel} in short as
\begin{equation}
\mathbf{o} = \left[\begin{array}{ccc} H_x  & H_a \end{array}\right]\left[ \begin{array}{c} \mathbf{x} \\ \mathbf{a} \end{array} \right] + \nu_G \label{Eq:Gps_meas_equation}
\end{equation}
where the de-correlated noise vector $\nu_G = (\tilde\nu_\eta, \tilde\nu_\xi, \tilde\nu_\zeta)^T$ is distributed as $\nu_G \sim [I_{3M-3}, \mathbf{0}]$.

\subsection{State Process Equation}
Referring \eqref{Eq:bias_dynamics}, and \eqref{Eq:scale_dynamics}, we have
\begin{align}
\mathbf{f}^v &= \tilde{\mathbf{f}}^v + S_f \tilde{\mathbf{f}}^v + \boldsymbol{\beta}_f \nonumber\\
\boldsymbol{\omega}^v_i &= \tilde{\boldsymbol{\omega}}^v_i + S_\omega \tilde{\boldsymbol{\omega}}^v_i + \boldsymbol{\beta}_\omega \nonumber
\end{align}
where $\tilde{\mathbf{f}}^v$ and $\tilde{\boldsymbol{\omega}}_i^v$ are the actual sensor readings of acceleration and angular rate, respectively. Plugging above two equations into the discrete version of \eqref{Eq:naveq}, we obtain the system process equation \eqref{Eq:SystemProcessEqu} where $\mathbf{x}_{t+1}$ and $\mathbf{x}_{t}$ is the state vector at time steps $t+1$ and $t$, respectively; $[\tilde{\mathbf{f}}^v]$ and $[\tilde{\boldsymbol{\omega}}_i^v]$ are matrices whose diagonal entries are $\tilde{\mathbf{f}}^v$ and $\tilde{\boldsymbol{\omega}}_i^v$, respectively;  $\Delta t$ is the duration between two consecutive two steps; $\mathbf{k}_r$, $\mathbf{k}_v$, and $\mathbf{k}_\omega$ are random vectors of zero-mean Gaussian distributions that models the un-modeled uncertainties (e.g., time jitters and errors from model parameters) in \eqref{Eq:naveq}; $\sigma_{\beta_f}$, $\sigma_{\beta_\omega}$, $\sigma_{\beta_S}$, and $\sigma_{S_\omega}$ are sensor error parameters defined in Table I;  random vectors $\mathbf{w}_{\beta_f}\sim \mathcal{N}(\mathbf{0},\mathbf{I}_3)$, $\mathbf{w}_{\beta_\omega}\sim \mathcal{N}(\mathbf{0},\mathbf{I}_3)$, $\mathbf{w}_{S_f}\sim \mathcal{N}(\mathbf{0},\mathbf{I}_3)$, and $\mathbf{w}_{S_\omega}\sim \mathcal{N}(\mathbf{0},\mathbf{I}_3)$; and $\mathbf{I}_3$ is a $3\times 3$ identity matrix.

\begin{table}[h]
\caption{Error model parameters for CrossBow Fiber optical gyro system FG700AB and speedometer. $\sigma_{\beta_f}$ and $\sigma_{\beta_\omega}$ are the random walk parameters for acceleration and angular rate, respectively. $\sigma_{S_f}$ and $\sigma_{S_\omega}$ are the scale factor parameters for acceleration and angular rate, respectively. $\sigma_{\beta_v}$ and $\sigma_{S_v}$ are the error parameters of speedometer. }
\centering
\begin{tabular}{|c|c|c|c|c|c|}
  \hline
  $\sigma_{\beta_f}$ & $\sigma_{\beta_\omega}$ & $\sigma_{S_f}$ & $\sigma_{S_\omega}$ & $\sigma_{\beta_v}$ & $\sigma_{S_v}$\\
  $\text{m/s/s}^{1/2}$ & $^\circ/\text{s}^{1/2}$ & & & $\text{m/s}^{1/2}$ &\\
  \hline
  0.0167 & 0.0067 & 0.01 & 0.01 & 0.05 & 0.01\\
  \hline
\end{tabular}
\end{table}

\section*{{}}
\begin{widetext}
\small
\begin{align}
\mathbf{x}_{t+1}&=
\left(
\begin{array}{ccccccc}
\mathbf{I}_3 & \mathbf{I}_3 \Delta t & \mathbf{0}_3 & \mathbf{0}_3 & \mathbf{0}_3 & \mathbf{0}_3 & \mathbf{0}_3\\
\mathbf{0}_3 & \mathbf{I}_3 - 2 \boldsymbol{\Omega}_i^n\Delta t & \mathbf{0}_3 & R_v^n\Delta t & \mathbf{0}_3 & R_v^n [\tilde{\mathbf{f}}^v]\Delta t & \mathbf{0}_3 \\
\mathbf{0}_3 & \mathbf{0}_3 & \mathbf{I}_3 & \mathbf{0}_3 & \mathbf{I}_3\Delta t & \mathbf{0}_3 & [\tilde{\boldsymbol{\omega}}_i^v]\Delta t \\
\mathbf{0}_3 & \mathbf{0}_3 & \mathbf{0}_3 & \mathbf{I}_3 & \mathbf{0}_3 & \mathbf{0}_3 & \mathbf{0}_3 \\
\mathbf{0}_3 & \mathbf{0}_3 & \mathbf{0}_3 & \mathbf{0}_3 & \mathbf{I}_3 & \mathbf{0}_3 & \mathbf{0}_3 \\
\mathbf{0}_3 & \mathbf{0}_3 & \mathbf{0}_3 & \mathbf{0}_3 & \mathbf{0}_3 & \mathbf{I}_3 & \mathbf{0}_3 \\
\mathbf{0}_3 & \mathbf{0}_3 & \mathbf{0}_3 & \mathbf{0}_3 & \mathbf{0}_3 & \mathbf{0}_3 & \mathbf{I}_3
\end{array}
\right)
\mathbf{x}_{t}
+ \mathbf{u}_t
+\left(
\begin{array}{c}
\mathbf{k}_r\\
\mathbf{k}_v\\
\mathbf{k}_\omega \\
\sigma_{\beta_f} \sqrt{\Delta t} \mathbf{w}_{\beta_f}\\
\sigma_{\beta_\omega} \sqrt{\Delta t} \mathbf{w}_{\beta_\omega}\\
\sigma_{S_f} \sqrt{\Delta t} \mathbf{w}_{S_f}\\
\sigma_{S_\omega} \sqrt{\Delta t} \mathbf{w}_{S_\omega}
\end{array}
\right)
\label{Eq:SystemProcessEqu} \\
\mathbf{u}_t &= \left(
\begin{array}{ccccccc}
\mathbf{0}_3 &
(-(\boldsymbol{\Omega}_e^n)^2\mathbf{o}_e^n + R_v^n \tilde{\mathbf{f}}^v + \mathbf{g}^n)^T\Delta t &
(\tilde{\boldsymbol{\omega}}_i^v - R_e^v \boldsymbol{\omega}_i^e)^T\Delta t&
\mathbf{0}_3 &
\mathbf{0}_3 &
\mathbf{0}_3 &
\mathbf{0}_3
\end{array}\right)^T \nonumber
\end{align}
\end{widetext}

\subsection{LVNS Velocity Constraint}
We further augment the state vector $\mathbf{x}$ with the error two parameters $\beta_v$ and $S_v$ (c.f., Section~\ref{sc:wheelencoderErr}) for vehicle velocity measurement. The process equations for $\beta_v$ and $S_v$ can be modeled similarly as in Eq.~\eqref{Eq:bias_dynamics} and \eqref{Eq:scale_dynamics}, respectively. We have
\begin{equation}
\left(\begin{array}{c}
\beta_v(t+1) \\
S_v(t+1)
\end{array}\right) =
\left(\begin{array}{c}
\beta_v(t) \\
S_v(t)
\end{array}\right)
+
\left(\begin{array}{c}
\sigma_{\beta_v} \sqrt{\Delta t}w_{\beta_v} \\
 \sigma_{S_v} \sqrt{\Delta t} w_{S_v}
\end{array}\right) \label{Eq:EncoderSpdErrModel}
\end{equation}
where $w_{\beta_v}\sim \mathcal{N}(0,1)$ and $w_{S_v} \sim \mathcal{N}(0,1)$ are Gaussian random variables;  $\sigma_{\beta_v}$ and $\sigma_{S_v}$ are the parameters of standard deviation specified in Table I.

Note that \eqref{Eq:SystemProcessEqu} and \eqref{Eq:EncoderSpdErrModel} can be combined and normalized to be
\begin{equation}
\mathbf{x}_{t+1} = F_t \mathbf{x}_{t} + \mathbf{u}_{t} + \mathbf{w}_t
\label{Eq:SystemProcessEqu2}
\end{equation}
where $\mathbf{w}_t \sim \mathcal{N}(\mathbf{0}, \mathbf{I})$.

Assuming the land vehicle does not slip and travels along the bore-sight of the vehicle (i.e., $x$-axis in $v$-frame), we have the vehicular velocity to be zero along the directions perpendicular to $x$-axis in $v$-frame, i.e., $\mathbf{v}^v = (v_H, 0, 0)^T$. Let the vehicle measurement $\tilde{\mathbf{v}}^v = (\tilde{v}_H, 0, 0)^T$ and $\tilde{v}_H$ is directly measured from the speedometer. Using \eqref{Eq:encoder_errmodel} we have
\begin{equation}
\tilde{\mathbf{v}}^v =  R_n^v \mathbf{v}^n
 + \left(\begin{array}{c} v_H \\
0 \\
0 \end{array}\right) S_v
 +
 \left(\begin{array}{c}
 1\\
 0 \\
 0
 \end{array}\right) \beta_v + \nu_V
 \label{Eq:LVNS_constraint}
\end{equation}
where $\nu_V \sim \mathcal{N}(\mathbf{0},\Sigma_{\nu})$ is the random vector denoting un-modeled disturbances.

Note that the above equation can be merged into the measurement equation \eqref{Eq:Gps_meas_equation}.

\subsection{Vehicle Trajectory Reconstruction} \label{sc:post-mission}
Here we discuss the post-mission data process. The objective is to obtain an optimal estimate for the vehicle trajectory and the GPS ambiguities $\mathbf{a}$, given all measurements available to us. By $\mathbf{x}_t$ and $\mathbf{o}_t$ we denote the augmented vehicle state and the measurement observation at discrete time step $t$, respectively. Let the entire vehicle trajectory as $\mathbf{X}_{1:\tau}=[\mathbf{x}_t]^T$ (up to epoch $\tau$), all the ambiguities as $\mathbf{a}$, and all t he measurements as $\mathbf{O}_{1:\tau} = [\mathbf{o}_t]^T$ (up to epoch $\tau$). Given $\mathbf{O}_{1:\tau}$, we use the Bayesian network (BN) to compute the maximum likelihood estimate of $\mathbf{X}_{1:\tau}$ and $\mathbf{a}$: \[
\max_{\mathbf{X}_{1:\tau}, \mathbf{a}} p(\mathbf{X}_{1:\tau}, \mathbf{a} \mid \mathbf{O}_{1:\tau})
\]

\begin{figure}[h]
    \centering
    \includegraphics[width=3.3in]{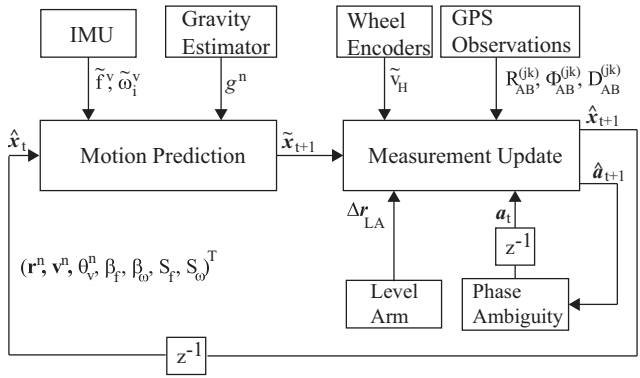}
    \caption{\protect\small  Block diagram of IMU and GPS integration for LVNS.}
    \label{FG:block-diagram}
\end{figure}

Fig.~\ref{FG:block-diagram} shows the block diagram of the proposed vehicular positioning system. The motion prediction module monitors the inputs from the IMU sensor, the gravity estimator, and the previous state vector $\hat{\mathbf{x}}_t$ using \eqref{Eq:SystemProcessEqu2}. The output from the motion prediction module $\tilde{\mathbf{x}}_{t+1}$ is updated based on the data from GPS observations (i.e., pseudo-range measurement $R^{(jk)}_{AB}$, phase measurement $\Phi^{(jk)}_{AB}$, and Doppler measurement $D^{(jk)}_{AB}$), the vehicle speed from speedometer (i.e., $\tilde{x}_H$), and the known level-arm $\Delta \mathbf{r}_{\text{LV}}$. The measurement update module computes the new estimate of the state $\hat{\mathbf{x}}_{t+1}$, and phase measurement ambiguity vector $\mathbf{a}$.

\begin{figure}[h]
    \centering
    \includegraphics[width=2.7in]{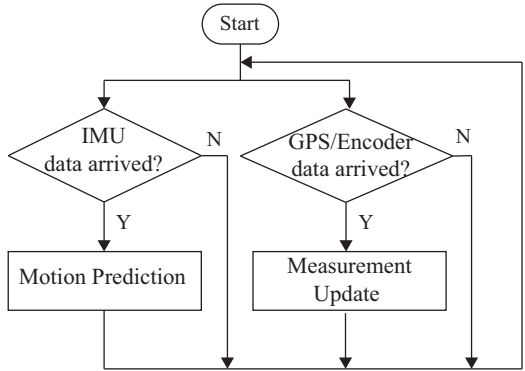}
    \caption{\protect\small Flowchart diagram of IMU and GPS integration for LVNS.}
    \label{FG:flow-chart}
\end{figure}

Fig.~\ref{FG:flow-chart} shows the flowchart of the proposed positioning system. Since the data refreshing rates between IMU and GPS/speedometer are different and not synchronized, we use the event driven structure to process the data. The motion prediction module is called whenever new data from IMU sensor is arrived. Measurement update using \eqref{Eq:Gps_meas_equation} and \eqref{Eq:LVNS_constraint} are triggered once  a new measurement from GPS and speedometer.

\section{Conclusions and Future Work} \label{SC:Conclusion}
We have described the implementation details of the positioning system that integrates GPS measurements (i.e., pseudo-range, carrier-phase and doppler), IMU measurements, and speedometer measurements. We derived the state process equation for motion prediction, the GPS measurement equation, and speedometer measurement equation. From these linearized equations, the techniques of extended Kalman filtering (EKF) or BN can be applied to jointly estimate the vehicle trajectory and the phase ambiguity.


\bibliographystyle{IEEEtranS}
\bibliography{../../gm_eci}
\begin{IEEEbiography}[{\includegraphics[width=1in,height=1.25in,clip,keepaspectratio]{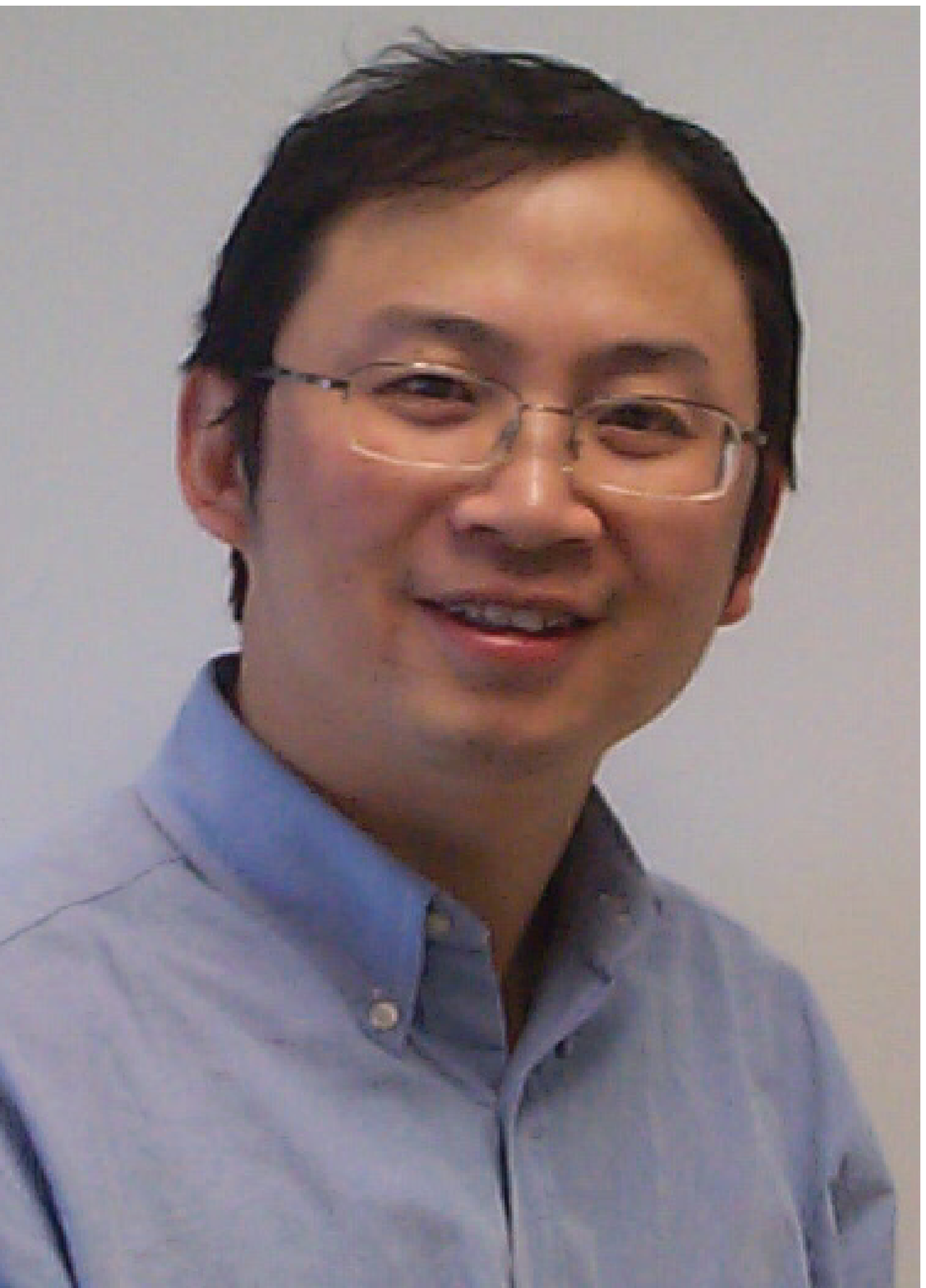}}]{Shuqing Zeng (M'03)}
received his PhD degree in Computer Science from the Michigan State University, East Lansing, Michigan, in 2004.

Since 2004, he has been with the Research and Development Center, General Motors Corporate, Warren, MI, where he currently holds the position of Senior Research Scientist. From 2005 to 2007, he served as the newsletter Editor of Autonomous Mental Development TC, IEEE Computational Intelligence Society. He is currently an Associate Editor of the International Journal of Humanoid Robotics. His research interests include computer vision, sensor fusion, autonomous driving, and active-safety applications on vehicle.

Dr. Zeng has served as a judge to the Intelligent Ground Vehicle Competition. He is a member of the Tartan Racing team who won the first place of the Defense Advanced Research Projects Agency Urban Challenge on November 3, 2007.
\end{IEEEbiography}

\end{document}